\documentclass{article}
\usepackage{ijcai24}

\usepackage{times}
\usepackage{soul}
\usepackage{url}
\usepackage[hidelinks]{hyperref}
\usepackage[utf8]{inputenc}
\usepackage[small]{caption}
\usepackage{graphicx}
\usepackage{amsmath}
\usepackage{amsthm}
\usepackage{booktabs}
\usepackage{algorithm}
\usepackage{algorithmic}
\usepackage[switch]{lineno}

\usepackage{bm}
\usepackage{amssymb}
\usepackage[table]{xcolor}
\usepackage{pgfplots}
\usepackage{subcaption}
\usepackage{tikz}
\usepackage{multirow}

\pgfplotsset{compat=newest}

\definecolor{mygray}{gray}{.9}
\definecolor{sota_blue}{HTML}{0071bc}
\makeatletter
\newcommand{\thickhline}{%
	\noalign {\ifnum 0=`}\fi \hrule height 1pt
	\futurelet \reserved@a \@xhline
}
\makeatother
\newcommand{\tablestyle}[2]{\setlength{\tabcolsep}{#1}\renewcommand{\arraystretch}{#2}\centering\footnotesize}

\title{Landmark Guided Active Exploration with State-specific Balance Coefficient}

\author{
Fei Cui$^{1}$
\and
Jiaojiao Fang$^{1}$\and
Mengke Yang$^{1}$\And
Guizhong Liu$^1$\thanks{Corresponding author}
\affiliations
$^1$Xi'an Jiaotong University\\
}

\begin{document}

\maketitle

\begin{abstract}
    Goal-conditioned hierarchical reinforcement learning (GCHRL) decomposes long-horizon tasks into sub-tasks through a hierarchical framework and it has demonstrated promising results across a variety of domains. However, the high-level policy’s action space is often excessively large, presenting a significant challenge to effective exploration and resulting in potentially inefficient training. In this paper, we design a measure of \textit{prospect} for subgoals by planning in the goal space based on the goal-conditioned value function. Building upon the measure of prospect, we propose a landmark-guided exploration strategy by integrating the measures of \textit{prospect} and \textit{novelty} which aims to guide the agent to explore efficiently and improve sample efficiency. In order to dynamically consider the impact of prospect and novelty on exploration, we introduce a state-specific balance coefficient to balance the significance of prospect and novelty. The experimental results demonstrate that our proposed exploration strategy significantly outperforms the baseline methods across multiple tasks.
\end{abstract}

\section{Introduction}
\label{sec:intro}
Deep reinforcement learning (DRL) is a powerful approach to sequential decision problems, such as video games \cite{schrittwieser2020mastering} and robot navigation \cite{dugas2021navrep,kastner2021connecting,li2022memonav}. DRL models these problems as partially observable Markov decision processes (POMDPs), and learns the optimal policies by maximizing the cumulative discounted reward. However, in many complex tasks, agents often struggle to collect sufficient high-reward trajectories. Hierarchical reinforcement learning (HRL) decomposes complex, long-horizon decision tasks into sub-tasks of different time scales and is a promising method for solving such long-horizon tasks. Goal-conditioned hierarchical reinforcement learning (GCHRL) \cite{dayan1992feudal,kulkarni2016hierarchical,nachum2018near,wang2022hierarchical} is a two-level hierarchical reinforcement learning paradigm, where the high-level policy decomposes the original task into a series of subgoals, and the low-level policy guides the agent to achieve these subgoals.
Effective subgoals are crucial for achieving good performance and efficiency in GCHRL. Selecting reasonable subgoals that capture the task's semantics provides meaningful guidance to low-level policy learning. Pre-defined subgoal space \cite{kim2021landmark,zhang2020generating} and task-specific subgoal representation space \cite{li2021learning,liactive,pere2018unsupervised,sukhbaatar2018learning} learned online can be employed to better represent the action space of the high-level policy. Pre-defined subgoals can quickly supervise the low-level policy learning via intrinsic rewards, while learned subgoal representations can be optimized for specific tasks. However, sampling actions in a large subgoal space can lead to inadequate exploration and inefficient training of the high-level policy.

To enhance agent's exploration, some approaches \cite{kim2021landmark,zhang2020generating} reduce the complexity of the high-level action space with adjacency constraints, encouraging exploration of reasonable states. HESS \cite{liactive} addresses non-stationarity in HRL with a exploration strategy considering \textit{novelty} and \textit{potential} measures for subgoals. 
Novelty promotes exploration of new states, and potential guides the agent towards expanding the explored area. Despite of its effectiveness, HESS does not account for the impact of the final goal on exploration, potentially limiting guidance to the most promising areas.


Our insight is that the agent not only needs to expand the exploration area but also needs to pay attention to the areas that are more likely to guide the agent to the final task goal. To this end, we design a \textit{prospect} measure for subgoals through landmark-based planning in the goal space.
This measure can reflect the likelihood of exploring in the direction of a subgoal leading the agent to states closer to achieving the final task goal. Considering the measure of \textit{prospect} and \textit{novelty} for subgoals, we propose a {\bf{L}}andmark-guided active {\bf{E}}xploration strategy with \textbf{S}tate-specific Balance \textbf{C}oefficient (\textbf{LESC}). The strategy incorporates the measure of \textit{prospect} to guide the agent to explore subgoals that are more likely to lead to the final task goal. 
Additionally, addressing the balance between the prospect and the novelty measures is crucial. When the prospect measure does not effectively reflect the guidance to the goal in the exploration, the agent should be guided by the novelty measure to explore new areas, aiming to overcome maze obstacles. Therefore, we designe a dynamic balance coefficient to enable the agent to receive guidance from landmarks while simultaneously maintaining the capability to explore new areas.
The proposed LESC strategy effectively balances the exploration-exploitation trade-off.

We compare the proposed method LESC with the state-of-the-art baselines in the Mujoco simulator \cite{todorov2012mujoco}. The experimental results demonstrate that LESC, which takes into account the guidance of the landmarks for exploration, outperforms the baseline methods. Additionally, we conduct ablation experiments to verify the roles of different components of LESC.


\section{Related Work}
\subsection{Hierarchical Reinforcement Learning}
Hierarchical Reinforcement Learning decomposes the original task into sub-tasks at different timescales using a hierarchical structure. The high-level policy communicates with the low-level policy through subgoals, and the signal passed from the high-level policy to the low-level policy can vary across different tasks, ranging from using discrete values for option \cite{duan2020hierarchical,fox2017multi,gregor2016variational} to employing pre-defined subgoal space \cite{kim2021landmark,nachum2018data,zhang2020generating} or subgoal representation space learned online \cite{ghoshlearning,li2021learning,liactive,pere2018unsupervised,sukhbaatar2018learning,vezhnevets2017feudal}. The use of discrete-valued options naturally reduces the complexity of the high-level action space, but the limited rules decrease the adaptability to complex tasks. On the other hand, learning subgoal representation often results in a high-dimensional action space, which can hinder the agent's ability to explore effectively. To explore effectively, HRAC \cite{zhang2020generating} restricts the high-level action space within the $k$-step adjacency area through adjacency constraints, while HIGL \cite{kim2021landmark} samples landmarks and restricts the actions of the high-level policy within the domain of the most urgent landmark.
In order to choose appropriate subgoals to guide exploration, HESS \cite{li2022hierarchical} proposes an exploration strategy by considering the measure of \textit{novelty} and \textit{potential} for subgoals but ignores the guidance of the task's final goal for exploration. In contrast, our approach plans a path of landmarks in the goal space and designs a measure of \textit{prospect} that considers the influence of the task's final goal, proposing a more efficient hierarchical exploration strategy.

\subsection{Subgoal Selection}
When deep reinforcement learning is employed to solve complex sequential decision tasks, selecting appropriate subgoals for the agent can be an effective strategy. Several studies \cite{chane2021goal,li2022hierarchical,nasiriany2019planning,pong2018temporal,zhang2021world} focus on learning goal-conditioned value functions and rationally design reward functions to make the value function reflect the reachability between two states. These approaches allow for planning in the goal space based on reachability, with states on the planned path selected as subgoals. 
After utilizing a value function to learn the reachability between states in the goal space, L3P \cite{zhang2021world} clusters the candidate states based on their reachability, with each cluster center representing a potential landmark. 
On the other hand, HIGL \cite{kim2021landmark} uses coverage-based sampling and novelty to sample landmarks from the replay buffer. To facilitate effective exploration with reasonable subgoals, HESS \cite{liactive} samples candidate landmarks in the neighborhood of the agent’s current state and selects subgoals based on the measures of \textit{novelty} and \textit{potential}. This active exploration strategy avoids introducing additional non-stationarity and speeds up the learning process of the hierarchical policy. Like previous works \cite{pong2018temporal,zhang2021world}, we also utilize a goal-conditioned value function to plan landmarks in the goal space. We design a measure of \textit{prospect} based on the landmarks. Considering the measures of \textit{novelty} and \textit{prospect} to select subgoals, we propose an active hierarchical exploration strategy.  

\section{Preliminaries}
Reinforcement learning (RL) formulates the sequential decision problem as a Markov Decision Process (MDP) \cite{sutton2018reinforcement}, defined as a quin tuple $M= <\mathcal{S}, \mathcal{A}, \mathcal{P}, r, \gamma>$
where $\mathcal{S}$ denotes the state space, $\mathcal{A}$ denotes the action space, $\mathcal{P}$ represents the state transition probability that reflects the dynamics of the environment, $r$ is the reward function typically designed by human experts for the task, and $\gamma \in [0,1)$ is a discount factor. A policy $\pi(a|s)$ maps a given state $s$ to a probability distribution over the action $a$. The objective of reinforcement learning is to learn a policy that maximizes the expected cumulative discounted reward $\mathbb{E}_{\pi} [\sum_{t=0}^{\infty} \gamma^t r_t]$, where $r_t$ is the immediate reward that the agent receives from the environment after taking action $a_t$ at state $s_t$. Reinforcement learning is mainly divided into two categories: the value-based methods \cite{wang2016dueling,van2016deep} and the policy gradient methods \cite{haarnoja2018soft,schulman2017proximal,wang2016dueling}. Value-based methods compute the state-action value function and choose actions greedily based on the computed value function. Policy gradient methods optimize the policy directly by the policy gradient computed on the value function. To encourage the agent to explore the state space, the Soft Actor-Critic (SAC) \cite{haarnoja2018soft} algorithm is adopted for both the high-level and the low-level policies in our experiments. In SAC, the standard value loss function is:
\begin{equation}
\begin{split}
    \mathcal{L}_Q(\theta )&=\mathbb{E}_{s_t, a_t,s_{t+1}\sim \mathcal{B},a_{t+1}\sim\pi_\psi} [\frac{1}{2}(Q_\theta (s_t, a_t)-(r(s_t, a_t)\\
    &+\gamma Q_\theta (s_{t+1}, a_{t+1})-\varsigma \textrm{log}(\pi_\psi (a_{t+1}|s_{t+1}))))^{2}]
\end{split}
\label{SAC value loss}
\end{equation}
where $\gamma$ is the discount factor, $\varsigma$ denotes the temperature coefficient, $\mathcal{B}$ is the replay buffer to store interactive trajectories, and $\pi_\psi$ represents the policy. The policy loss function is:
\begin{equation}
\begin{split}
   \mathcal{L}_\pi (\psi  )=\mathbb{E}_{s_t\sim\mathcal{B}, a_t\sim\pi_{\psi}}[\textrm{log}\pi_\psi(a_t|s_t)-\frac{1}{\varsigma}Q_\theta(s_t,a_t)]
\end{split}
\label{SAC POLICY loss}
\end{equation}

\begin{figure}[!t]
  \begin{center}
     \includegraphics[width=7cm]{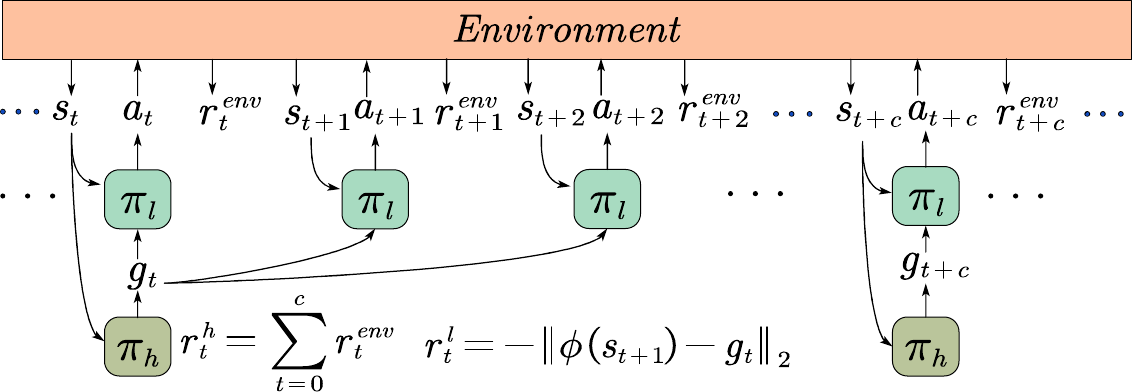}
  \end{center}
     \caption{The framework of GCHRL. Here $\phi$ is the subgoal representation function that maps the state to the goal space. The hierarchical framework consists of a high-level policy and a low-level policy. The reward of the high-level policy is a sum of c (the low-level policy length) external rewards, while the reward of the low-level policy is the negative distance between the state and subgoal in the goal space.}
  \label{fig:GCHRL}
\end{figure}

Goal-conditioned hierarchical reinforcement learning (GCHRL) models long-horizon decision tasks as a goal-conditioned Markov decision process, $M= <\mathcal{S}, \mathcal{G}, \mathcal{A}, \mathcal{P}, r, \gamma>$, where $\mathcal{G}$ represents the goal space. As illustrated in Figure \ref{fig:GCHRL}, GCHRL is a two-level framework. The high-level policy $\pi_h(g|s)$ operates at a lower temporal resolution, sampling a high-level action every c time steps (i.e., $t\equiv 0$ (\textrm{mod} $c$). When $t\not\equiv 0 $ (\textrm{mod} $c$), The subgoal generated by the high-level policy last time serve as the goal for the low-level policy. Given the current state $s_t$ and subgoal $g_t$, the low-level policy $\pi_l(a_t|s_t,g_t)$ produces a low-level action that is executed by the agent to interact with the environment. During training, since the objective of the low-level policy is to enable the agent to achieve the subgoals specified by the high-level policy, the low-level policy is optimized using the intrinsic reward $-\left \| \phi(s_{t+1})-g_t \right \|_2 $ computed from the subgoal, where $\phi$ is the subgoal representation function that maps the state to
the goal space. Similar to the standard reinforcement learning paradigm, the learning objective of the hierarchical structure is still to enable the agent to interact efficiently with the external environment, so the reward of the high-level policy is defined as the sum of $c$ external rewards $\sum_{t=0}^{c}r_t^{env}$ after executing a high-level action, here $r_t^{env}$ is the immediate reward obtained from the environment at time step $t$.

In GCHRL, the high-level policy and the low-level policy can be trained simultaneously in an end-to-end manner. The high-level policy provides real-time subgoals to the low-level policy and guides its learning through intrinsic rewards. The dynamic low-level policy learned online influences the stationarity of the high-level state transitions. Therefore, a well-designed hierarchical exploration strategy can significantly enhance the learning efficiency of the hierarchical policy.

\section{Method}

\subsection{Measures for Subgoals}
\begin{figure*}[t]
  \begin{center}
     \includegraphics[width=17cm]{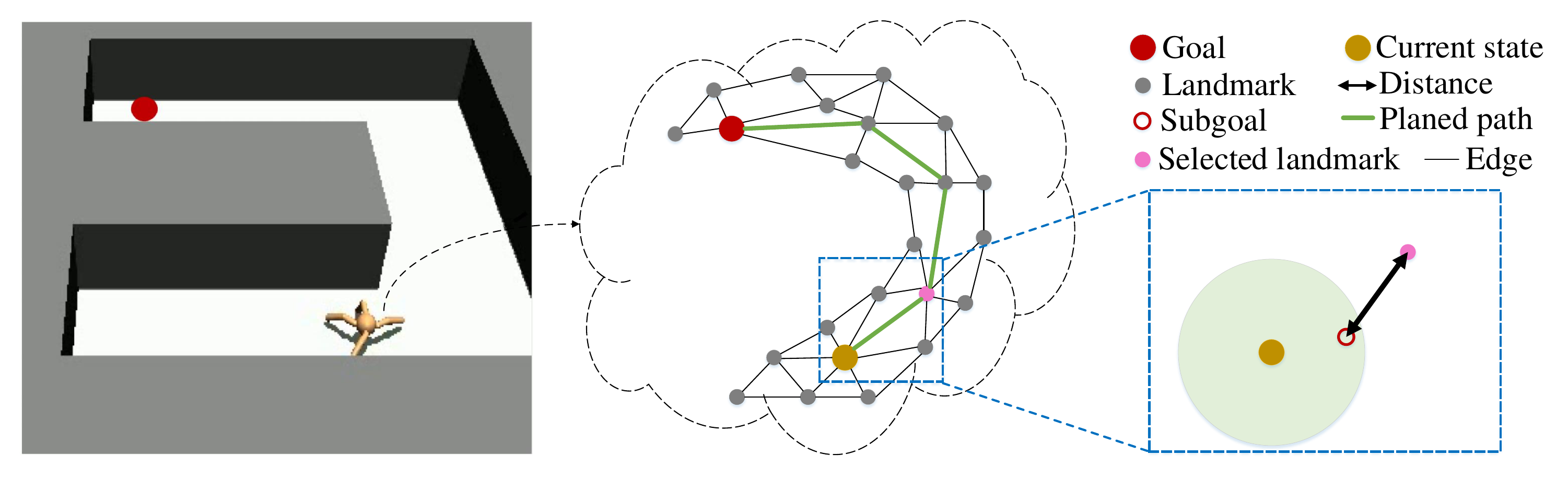}
  \end{center}
     \caption{landmark selection  and \textit{prospect} calculation process. The calculation of prospect involves four stages: 1) {\bf Sampling}: An adequate number of sample points are randomly selected from the state space. Then, the FPS algorithm is employed to sample $n_{cov}$ landmark points. 2) {\bf Building a graph}: The sampled landmarks, current position, and goal are used as nodes to build a graph. The edges of the graph represent the reachability between two nodes. 3) {\bf Path planing}: Using the shortest path planning algorithm, a feasible path from the current position to the goal is determined based on the constructed graph. 4) {\bf Calculation}: Landmark is sampled along the trajectory and selected as $l_{sel}$. The prospect of the subgoals (within the neighborhood of the current position) is then calculated based on the selected landmark.}
  \label{fig:landmark selection and prospect calculation process}
\end{figure*}

In GCHRL, effective exploration is crucial. Previous count-based methods \cite{kim2021landmark,machado2020count} rely on visit counts for subgoal's novelty measure, which may not consistently guide the agent to promising areas. The state-of-the-art approach \cite{liactive} introduces \textit{potential} as a measure for subgoals, aiming to guide the agent effectively. Our insight is that reasonable subgoals should not only expand exploration but also guide the agent to explore regions that are likely to lead to the ultimate goal. To address this, we introduce the measure of \textit{prospect} for subgoals, reflecting their positivity towards achieving the final goal.


The prospect measure requires specifying the exploration direction for the agent. Like prior works \cite{kim2021landmark,zhang2021world}, we choose to perform landmark-based planning in the goal space. We aim to sample landmarks that cover a wide range of the goal space. To achieve this, the Farthest Point Sampling (FPS) \cite{arthur2007k} algorithm is employed to sample $n_{cov}$ landmarks from buffer $\mathcal{B}$. Starting with an initial candidate set, the farthest landmark is iteratively added to the sampled landmark set until a sufficient number of landmarks are sampled, where the distance between two landmarks is measured by the Euclidean distance in the goal space. This sampling process ensures the sampled landmarks cover the entire goal space. To search for a promising path from the current state to the goal, we build a graph consisting of the current state, the goal, and the sampled landmarks as nodes; like previous works \cite{kim2021landmark,nasiriany2019planning,pong2018temporal,zhang2021world} and \cite{huang2019mapping}, the edges between nodes are weighted based on the goal-conditioned value function $V(s_1, \phi(s_2))$, where $\phi(s)$ is the subgoal representation function
that maps the state to the goal space. This value function reflects the reachability between state $s_0$ and the goal $g$ in the goal space, defined as:
\begin{equation}
\tablestyle{5cm}{1}\begin{split}
   V(s_0,g)&=\mathbb{E}_{\pi_{l}}[ {\textstyle \sum_{t=0}^{\infty }}\gamma^t \mathcal{R}(s_t,a_t,g)] \\
   \mathcal{R}(s_t,a_t,g)&=\begin{cases} \ 0 \ \ \left \| \phi(s_{t+1})-g \right \|_2< \delta    \\ \  -1 \ \ \ \   otherwise  \end{cases} 
\end{split}
\label{defineV(s,g)}
\end{equation}
where $\delta$ is the threshold for reaching the goal. Once the graph is built, a feasible path from the current state to the goal can be found through a shortest path algorithm, and the landmark $l_{sel}$ closest to the current state is selected as the region the agent should explore. If no feasible path is available, the goal $g$ is selected as $l_{sel}$. The process of landmark selection is described in Algorithm \ref{alg:alg1}.

\begin{algorithm}[H]
\caption{Landmark Selection}
\label{alg:alg1}
\begin{algorithmic}[1]
\STATE \textbf{Input: } current state $s_t$, goal $g$, clip value $\tau$, $\phi(s)$
\STATE \hspace{1.2cm}and goal-conditioned value function $V(s, g)$
\STATE \textbf{Output: } selected lanmark $l_{sel}$
\STATE Initial $n$ transitions $T = (s, a, s')$ from buffer $\mathcal{B}$
\STATE $X\gets \textbf{FPS} (S = \left \{ s\parallel (s, a,s' )\in T  \right \} )\cup \left \{ g \right \}$
\STATE $W _{i,j} \gets \infty $
\STATE \textbf{for } $\forall (n_i, n_j) \in X\times X$ \textbf{do}
\STATE \hspace{0.5cm} $w_{i,j} \gets  [-V(n_i, \phi(n_j))] $
\STATE \hspace{0.5cm} \textbf{if} $w_{i,j} \le \tau $ \textbf{then}
\STATE \hspace{1.0cm} $W_{i,j} = w_{i,j}$
\STATE \hspace{0.5cm} \textbf{end if}
\STATE \textbf{end for}
\STATE $Trajectory \ K \gets  \textbf{Shortest Path Planing}(X, W)  $
\STATE \textbf{if} $K$ is None \textbf{then}
\STATE \hspace{0.5cm} $l_{sel} = g$
\STATE \textbf{else}
\STATE \hspace{0.5cm} $k_{sel} \gets \text{argmin}_{k_i\in K} -V(s_t, \phi(k_i))$
\STATE \hspace{0.5cm} $l_{sel} = \phi(k_{sel})$
\STATE \textbf{return } $l_{sel}$
\end{algorithmic}
\label{alg1}
\end{algorithm}

In practice, before training the hierarchical policy, we allow the agent start at random positions and sample goals to collect trajectories stored in buffer $\mathcal{B}$. These trajectories are used to pre-train the goal-conditioned value function $V(s,g)$ in a bootstrapping way \cite{huang2019mapping} and facilitate performing furthest point sampling at the beginning of the training. For the current state $s_t$, after selecting landmark $l_{sel}$ with Algorithm \ref{alg:alg1}, the exploration strategy considers selecting a subgoal $g_t$ near the current state $s_t$, the \textit{prospect} of the subgoal $g_t$ is defined as:
\begin{equation}
\begin{split}
   P(g_t) = -\left \| g_t-l_{sel} \right \|_2
\end{split}
\label{prospect define}
\end{equation}

The landmark selection and prospect calculation process are depicted in Figure \ref{fig:landmark selection and prospect calculation process}. The \textit{prospect} measure takes into account the influence of the goal on the subgoal selection by imaging a feasible path to the goal in the goal space. In contrast to previous works, the prospect measure not only encourages the agent to explore unexplored region but also guides the agent to explore regions that have a positive impact on achieving the task's goal. To maintain the agent's ability to explore new areas, like previous works \cite{machado2020count,kim2021landmark,liactive}, we also consider the \textit{novelty} measure for subgoals. We employ the method of random network distillation (RND) \cite{burda2018exploration} to assess the novelty of the states. RND initializes a neural network $f$ with fixed parameter $\bar{\vartheta} $ and a trainable network  $\hat{f}$ parameterized by $\vartheta$, which is trained using mean squared error.
\begin{equation}
\begin{split}
   \mathcal{L}_{RND}(\vartheta)=\mathbb{E}_{s\sim \mathcal{B}}\left \| f(s;\bar{\vartheta} ) -\hat{f}(s;\vartheta) \right \|_2 
\end{split}
\label{RND loss}
\end{equation}
For states frequently visited by the agent, training is more thorough, resulting in lower $\mathcal{L}_{RND}$. The novelty measure should incentivize the exploration of remote and uncharted areas. Therefore, the novelty of state $s_i$ is defined as the expected cumulative RND loss of current state and future states:
\begin{equation}
\tablestyle{5.5cm}{1}\begin{split}
   N(s_i)= \mathbb{E}_{s\sim\mathcal{B} }   {\textstyle \sum_{j=0}^{\left \lfloor (T-i)/c \right \rfloor }} \gamma ^j  \left \| f(s_{i+jc};\bar{\vartheta} ) -\hat{f}(s_{i+jc};\vartheta) \right \|_2 
\end{split}
\label{Novelty define}
\end{equation}
where $c$ is the low-level policy length, $\gamma$ is the discount factor.

\subsection{State-specific Balance Coefficient}
We design a measure of \textit{prospect} that efficiently guides the agent to explore the areas that have a positive impact on reaching the task goal. When assessing the impact of subgoals on exploration performance, we integrate the normalized measure of \textit{prospect} $\tilde{P}$ and \textit{novelty} $\tilde{N}$, expressed as the composite measure $(1-\alpha ) \tilde{P}+\alpha \tilde{N}$, where $\alpha$ represents the balance coefficient. During the execution of the exploration strategy, it is not feasible to plan a path to the final goal when the sampled $n_{cov}$ landmarks do not adequately cover the goal space or the estimation of the goal-conditioned value is inaccurate. In instances where the path-planning fails, the selected landmark $l_{sel}$ according to Algorithm \ref{alg:alg1} may not be reachable from the current state. This inaccessibility can be reflected by the goal-conditioned value function, i.e., $V(s_t,l_{sel})<-\tau $, here $\tau$ is the clip value for landmark selection. When the selected landmark $l_{sel}$ is unreachable, the exploration strategy should focus on the agent's ability to explore new areas, as measured by the novelty measure. Conversely, when the selected landmark is reachable, the agent should prioritize the guidance provided by the landmark $l_{sel}$. Therefore, we introduced a state-specific balance coefficient $\alpha$ to dynamically consider the influence of prospect and novelty measures on the exploration strategy. The state-specific balance coefficient $\alpha$ is defined as follows: 
\begin{equation}
\begin{split}
   \alpha = \begin{cases}\frac{-V(s_t,l_{sel})}{\tau } \  \ V(s_t,l_{sel}) \ge -\tau\\ \ \ \ \ \ \ \ 1 \ \ \ \ \ \ \ \ \ \ otherwise  \end{cases}
\end{split}
\label{state_specific alpha}
\end{equation}
When $V(s_t,l_{sel}) < -\tau$, signifying a lower accessibility of the landmark $l_{sel}$ with respect to the current state $s_t$, the balance coefficient $\alpha$ is set to 1, the evaluation of subgoals considers only the novelty measure. As the guidance of prospect becomes more pronounced, the balance coefficient progressively diminishes, indicating the increasing dominance of prospect in the exploration process.

\subsection{Hierarchical Exploration Strategy}
Blindly exploring unknown region may lead to accumulation of ineffective experiences.
To ensure that the subgoals generated by the exploration strategy are both promising and reachable from current state, we sample a candidate subgoal set $\mathcal{C}$ within the radius ($r_g$) neighborhood of the current state $s$. We then select the optimal subgoal by combining the prospect measure and the novelty measure, specified as follows:
\begin{equation}
\tablestyle{6cm}{1}\begin{split}
    g_t=\mathop{\mathrm{argmax}}\limits_{\phi(s)\in \mathcal{C} } (1-\alpha)\tilde{P}(\phi(s))+\alpha \tilde{N}(s)
\end{split}
\label{explorarion strategy}
\end{equation}
where $\tilde{N}$ and $\tilde{P}$ is the normalized novelty and prospect measure within the range [0,1]. $\mathcal{C}$ is the candidate subgoal set, $\phi$ is the subgoal representation function learned online, and $\alpha$ is the balance coefficient. 

Algorithm \ref{alg:alg2} describes our active hierarchical exploration strategy. In LESSON \cite{li2021learning}, it has been demonstrated that the triplet loss based on slow features can capture the relative positional relationships in the state space. Therefore, following LESSON, we train the subgoal representation function $\phi$ using the triplet loss, as follows:
\begin{equation}
\tablestyle{7cm}{1}\begin{split}
    \mathcal{L}_{\phi}=\mathbb{E}_{(s_t,s_{t+1},s_{t+c})\sim\mathcal{B} } [\left \| \phi(s_t)-\phi(s_{t+1}) \right \|_2\\ + max(0, \zeta -\left \| \phi(s_t)-\phi(s_{t+c}) \right \|_2 )]
\end{split}
\label{triplet loss}
\end{equation}
where $\mathcal{B}$ is the replay buffer, $\zeta$ is the margin parameter of triplet loss, $c$ is the low level policy length of the hierarchical framework.

\begin{algorithm}[H]
\caption{LESC algorithm}
\label{alg:alg2}
\begin{algorithmic}[1]
\STATE \textbf{Initialize} $p=0.7$, buffer size $1e6$
\STATE \textbf{Initialize: }$\pi_h(g|s),\pi_l(a|s, g)$, $\hat{f}(s)$, $\phi(s)$
\STATE \textbf{for }$i=1..episodeNum$ \textbf{do}
\STATE \hspace{0.5cm} \textbf{for } $t=0..T-1$ \textbf{do}
\STATE \hspace{1.0cm} \textbf{if} $t\equiv 0$ (\textrm{mod} $c$) \textbf{then}
\STATE \hspace{1.5cm} Uniformly sample $x$ in the range (0,1)
\STATE \hspace{1.5cm} \textbf{if} $x<p$ \textbf{then}
\STATE \hspace{2.0cm} Select landmark $l_{sel}$ with Algorithm \ref{alg:alg1}
\STATE \hspace{2.0cm} Generate a candidate set $\mathcal{C}$ containing $M$
\STATE \hspace{2.0cm} candidate subgoals
\STATE \hspace{2.0cm} Calculate the \textit{prospect} and \textit{novelty} of
\STATE \hspace{2.0cm} subgoals by Eq.\ref{prospect define} and Eq.\ref{Novelty define}
\STATE \hspace{2.0cm} Calculate balance coefficient $\alpha$ by Eq.\ref{state_specific alpha}
\STATE \hspace{2.0cm} Select subgoal $g_t$ by Eq.\ref{explorarion strategy}
\STATE \hspace{1.5cm} \textbf{else}
\STATE \hspace{2.0cm} Execute $g_t \sim \pi_h(\cdot| s_t)$
\STATE \hspace{1.5cm} Update $\pi_h$
\STATE \hspace{1.0cm} \textbf{else}
\STATE \hspace{1.5cm} $g_t=g_{t-1}$
\STATE \hspace{1.0cm} \textbf{end if}
\STATE \hspace{1.0cm} Execute $a_t\sim \pi(\cdot |s_{t},g_t)$
\STATE \hspace{1.0cm} Store experiences in replay buffer $\mathcal{B}$
\STATE \hspace{1.0cm} Update $\pi_l$ by Eq.\ref{SAC value loss} and Eq.\ref{SAC POLICY loss}
\STATE \hspace{1.0cm} Update $f$ by Eq.\ref{RND loss}
\STATE \hspace{0.5cm} \textbf{end for} 
\STATE \hspace{0.5cm} \textbf{if} $i\equiv 0$ (\textrm{mod} $I$) \textbf{then}
\STATE \hspace{1.0cm} Update $\phi$ using the triplet loss defined by Eq.\ref{triplet loss}
\STATE \hspace{0.5cm} \textbf{end if}
\STATE \textbf{end for}
\STATE \textbf{return: }$\pi_h, \pi_l, \hat{f}, \phi$
\end{algorithmic}
\label{alg2}
\end{algorithm}

\begin{figure*}[t]
  \begin{center}
     \includegraphics[width=\linewidth]{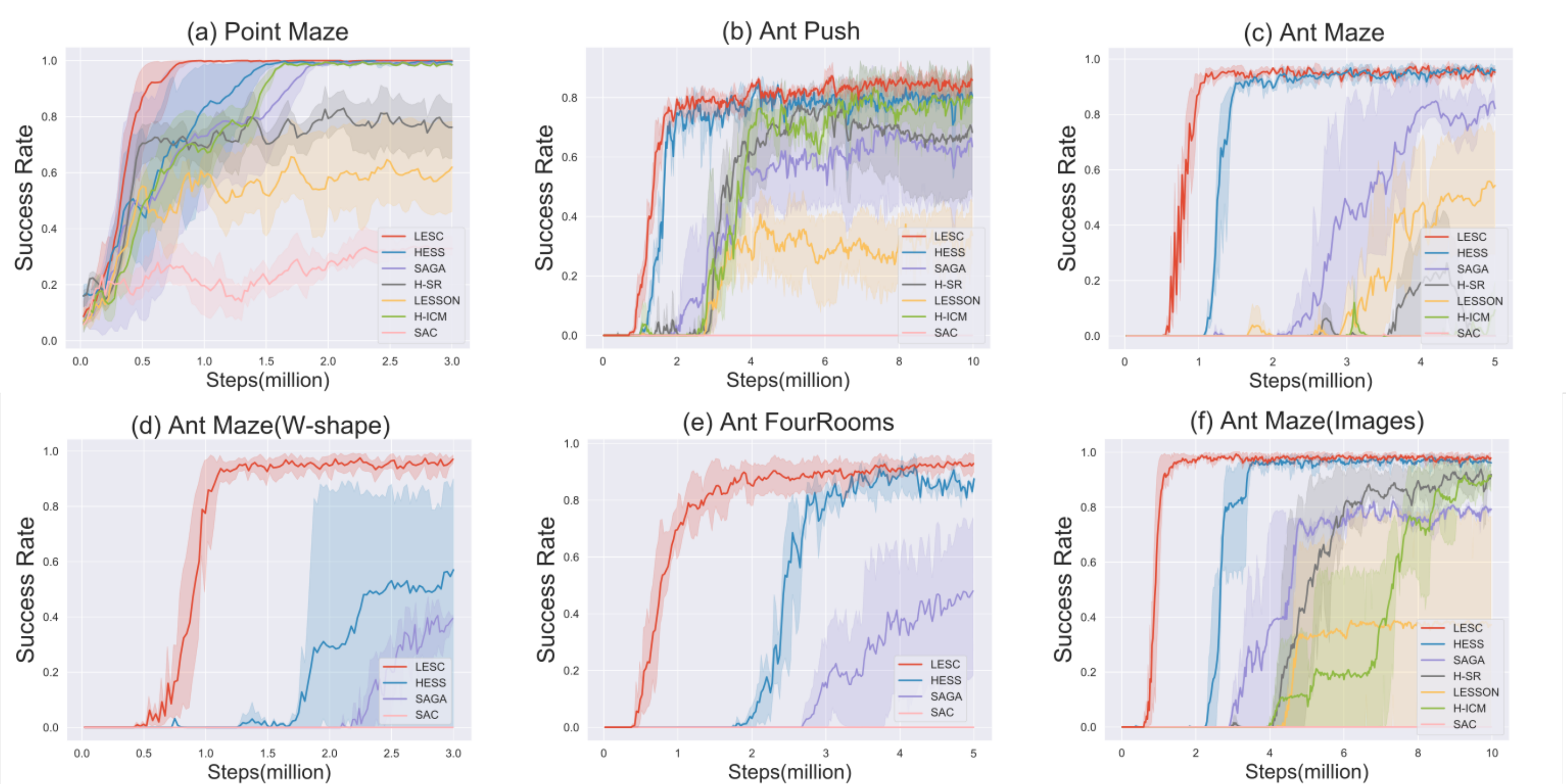}
  \end{center}
     \caption{Learning curves of LESC and baselines. (a) Point Maze. (b) Ant Push. (c) Ant Maze. (d) Ant Maze (W-shape). (e) Ant FourRooms. (f) Ant Maze (Images). The x-axis represents the training time steps, while the y-axis represents the average success rate over 50 episodes. The experiments are evaluated for each algorithm using five different random seeds. The shaded area represents the $95\%$ confidence interval.}
  \label{fig:Learning curves of LESC}
\end{figure*}


\section{Experiments}

\subsection{Experimental Setup}
We conduct experiments on several long-horizon decision tasks based on the Mujoco simulator, including Point (or Ant) Maze, Ant Maze (W-shape), Ant Push and Ant FourRooms. All the environments feature sparse rewards, intensifying the challenges and facilitating the assessment of the agent's exploration capabilities. 
The details of the environments, network architecture and hyper-parameters are provided in Supplementary Section A.

\begin{figure}[!t]
  \begin{center}
     \includegraphics[width=\linewidth]{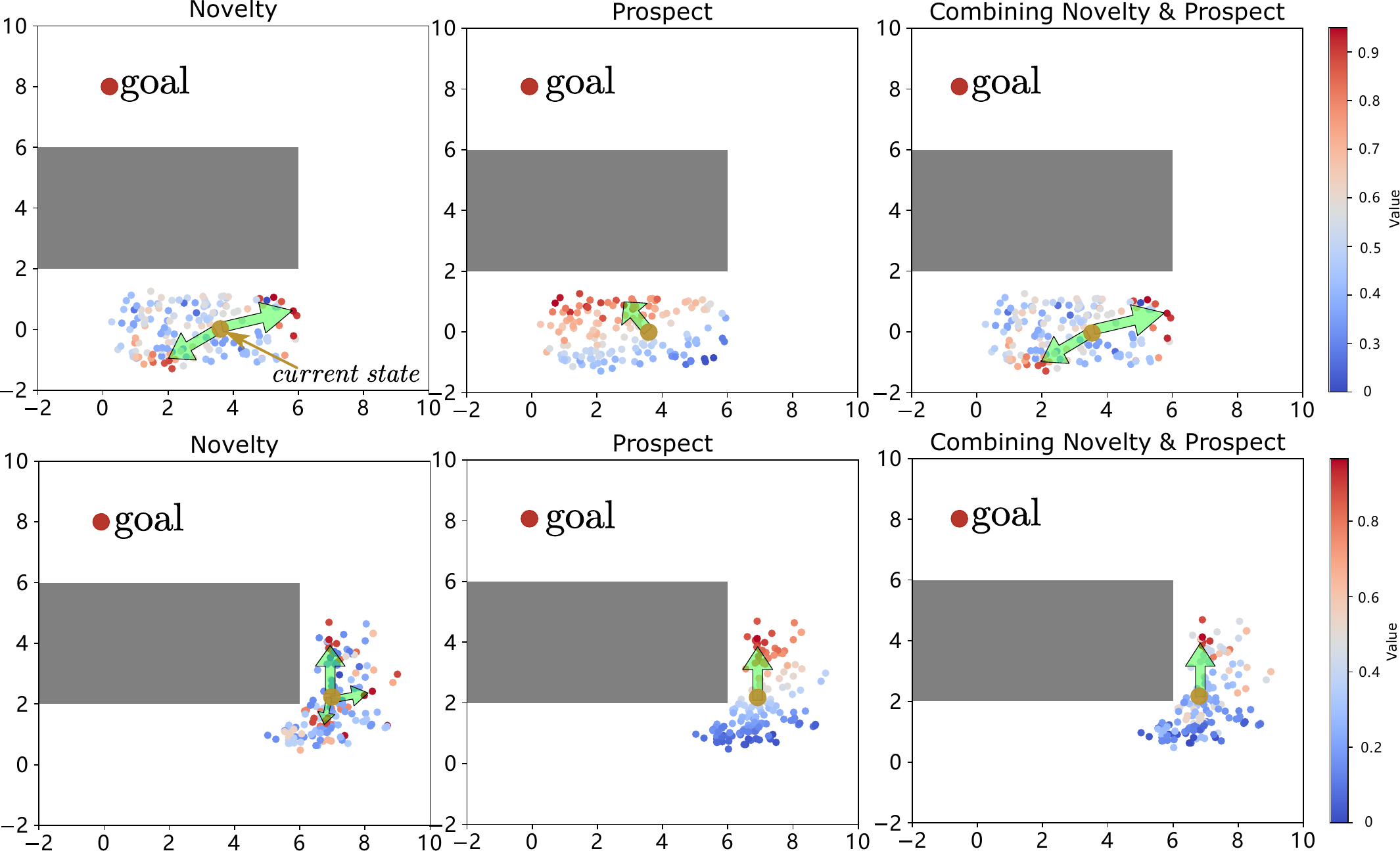}
  \end{center}
     \caption{The visualization of subgoal measures in the AntMaze task at time step 150000 (\textbf{up}) and time step 300000 (\textbf{down}). The circular markers represent the candidate subgoal set sampled by the agent. The color intensity of the markers, ranging from red to blue, indicates the corresponding measure values.}
  \label{fig:The visualization of subgoal measures}
\end{figure}

\begin{figure*}[t]
  \begin{center}
     \includegraphics[width=7in]{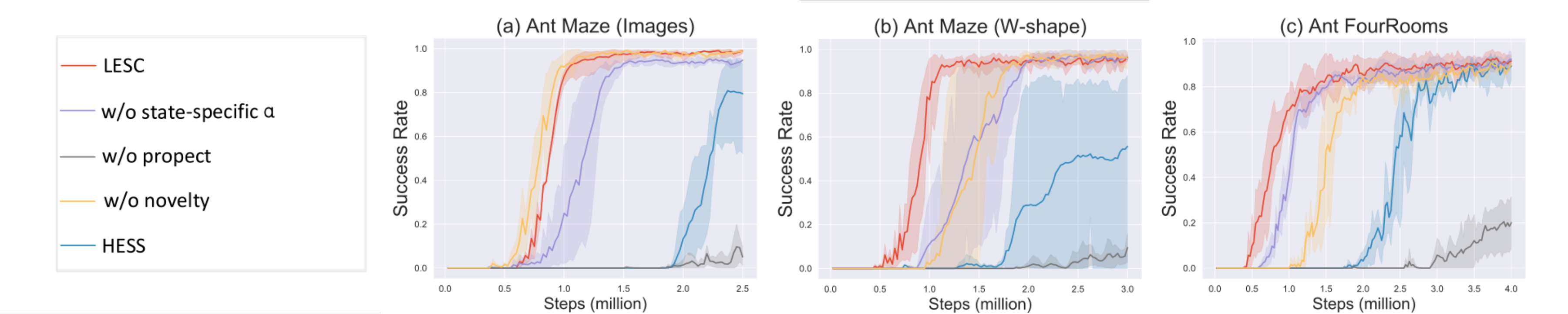}
  \end{center}
     \caption{Ablation studies on the components of LESC. 
    The experiments are evaluated using five different random seeds.}
  \label{fig:Ablation studies on the components}
\end{figure*}
\begin{figure*}[t]
  \begin{center}
     \includegraphics[width=7in]{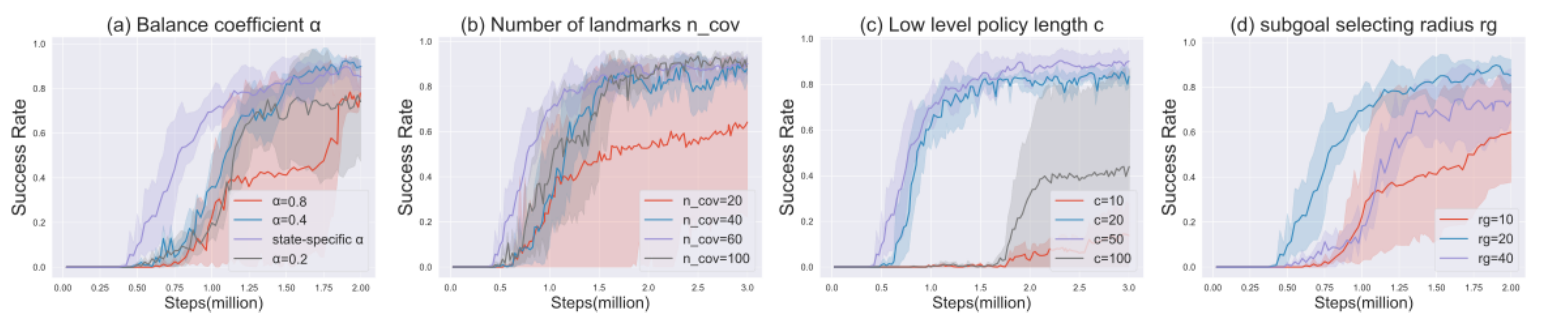}
  \end{center}
     \caption{Ablation studies on the hyper-parameters. All the ablation experiments are conducted specifically on the Ant FourRooms task.}
  \label{fig:Ablation studies on the hyper}
\end{figure*}
\subsection{Comparative Analysis}
We compare our method with several HRL methods, including SAGA \cite{wang2023state}, HESS \cite{liactive}, LESSON \cite{li2021learning}, H-SR \cite{machado2020count}, H-ICM \cite{pathak2017curiosity}, and SAC \cite{haarnoja2018soft}. SAGA introduces adversarial learning into GCHRL to alleviate the non-stationarity of high-level policy. HESS introduces \textit{potential} measure for subgoals. LESSON employs triplet loss for subgoal representation learning. H-SR utilizes count-based exploration with state visit counting. H-ICM uses curiosity as intrinsic rewards for exploration. SAC is a reinforcement learning algorithm based on the actor-critic mechanism. 

The experimental results depicted in Figure \ref{fig:Learning curves of LESC} demonstrate that, our proposed method LESC outperforms all the baseline methods in handling hard-exploration tasks. The superiority of LESC can be attributed to its comprehensive consideration of both the \textit{novelty} and the \textit{prospect} measure in the exploration strategy. 
Unlike the potential-based exploration approach HESS, which only focuses on expanding the exploration area, LESC effectively guides the agent towards regions that have a positive impact on reaching the task's goal. Additionally, we find that, in more complex tasks such as Ant FourRooms and Ant Maze (W-shape), which place greater demands on the exploration capability of the agent, the advantage of LESC in terms of sample efficiency becomes more pronounced. It can be observed that, H-ICM and H-SR fall short compared to our proposed method. 
H-ICM's performance depends on the learning error of the dynamic model, the instability of the intrinsic rewards restricts the learning of the high-level policy. H-SR calculates state visit counts using the $L_1$ norm of the successor representation, but relying solely on the successor representation, proves inadequate in promoting effective exploration. The experimental results demonstrate that, SAC exhibits the poorest performance when compared to the goal-conditioned hierarchical framework LESSON and SAGA. This finding serves as a strong evidence supporting the effectiveness of GCHRL.

\subsection{Qualitative Analysis on Measures for Subgoals}
To assess the impact of subgoal measures on subgoal selection, we visualize the \textit{prospect} and \textit{novelty} measures in the AntMaze task. As shown in Figure \ref{fig:The visualization of subgoal measures}, subgoals with high \textit{novelty} are distributed throughout the candidate subgoal set, indicating that the agent can explore in various directions to expand the explored area. At time step 150,000, the prospect measure fails to accurately guide the agent towards the path to the goal. Owing to state-specific balance coefficients, our exploration strategy, when combining novelty and prospect, does not blindly explore regions with high prospect but rather prioritizes novelty measures more. This ensures the agent's capability to explore new states when the prospect measure becomes ineffective. At time step 300,000, the prospect measure guides the agent towards the goal effectively. At this point, the state-specific balance coefficient prioritizes the guidance of the prospect measure, focusing on regions contributing significantly to task completion. 

\subsection{Ablation Studies}
\noindent{\bf Ablative analysis of various components}: 
To assess the significance of different components of LESC, we conduct an ablation study on the exploration strategy, evaluating performance under various settings: (\textbf{i}) the original LESC, (\textbf{ii}) LESC without state-specific balance coefficient $\alpha$, equal consideration of prospect and novelty (set $\alpha$ to 0.5), (\textbf{iii}) LESC without prospect, considering only novelty measure (set $\alpha$ to 1), (\textbf{iv}) LESC without novelty, considering only prospect measure (set $\alpha$ to 0), and (\textbf{v}) HESS, the state-of-the-art exploration baseline.
We conduct experiments in three different tasks: AntMaze (Images), AntMaze (W-shape), and AntFourRooms. The experimental results are depicted in Figure \ref{fig:Ablation studies on the components}. 
In the experimental setting without prospect, the algorithm performs poorly across multiple complex tasks. This emphasizes the crucial role of the prospect measure in the exploration process. In the simple Ant Maze, disregarding novelty led to the best performance, while in complex environments like Ant Maze (W-shape) and Ant FourRooms, neglecting novelty led to a performance decline. This indicates that in straightforward environments, computing the prospect measure consistently aids correct pathfinding towards the task's goal. However, as environmental complexity rises, the ability to plan a path to the goal is not always guaranteed. In such cases, the novelty measure becomes crucial to ensure the exploration capabilities of the agent. LESC demonstrates better performance compared to the setting without state-specific $\alpha$, especially in the AntMaze (W-shape) task. LESC dynamically considers the importance of prospect measure and novelty measure. When planning landmarks are unavailable, the consideration of novelty measure encourages the agent to explore new areas to overcome the maze obstacle. Conversely, when the landmark $l_{sel}$ is reachable, exploration of regions with high prospect is prioritized to facilitate the rapid collection of valuable interactive trajectories. These results highlights importance of dynamically considering prospect measure and novelty measure for addressing long-horizen tasks.

\noindent{\bf Ablative analysis on hyper-parameters}: 
Figure \ref{fig:Ablation studies on the hyper} presents ablation studies on several hyper-parameters. Ablation studies on additional parameters are provided in Supplementary Section B.

{\bf Balance coefficient $\alpha$}: In our proposed exploration strategy LESC, the balance coefficient $\alpha$ is used to balance the importance of the novelty and the prospect measure. In complex scenarios, choosing an appropriate balance coefficient is crucial. From Figure \ref{fig:Ablation studies on the hyper}(a), it can be observed that excessively large or small balance coefficients can result in a decline in performance. For all the experiments in Section 5.2, we choose a state-specific $\alpha$. 

{\bf Number of landmark samples $n_{cov}$}: LESC utilizes the FPS algorithm to sample $n_{cov}$ landmarks. It then selects the nearest landmark on the planned path as $l_{sel}$ (selected landmark). $n_{cov}$ influences the efficiency of path planning. Too small $n_{cov}$ may result in the inability to plan a feasible path to the goal. While too large $n_{cov}$ can lead to a burden on the shortest path planning and may result in the selected landmark $l_{sel}$ being too close to the current state, limiting the guidance to explore the most promising regions.
In the experiments conducted in Section 5.2,  for Ant FourRooms task, we set $n_{cov}$ to 60, to ensure comprehensive coverage of the environment. For Ant Maze (W-shape) task, $n_{cov}$ is set to 40. For other tasks, $n_{cov}$ is set to 20.

{\bf Low-level policy length $c$}: HRL decomposes long-horizon tasks into sub-tasks with finite horizon, denoted as $c$. Selecting an appropriate value for $c$ is advantageous as it allows for proper allocation of task difficulty between the high-level policy and low-level policy. Selecting an inappropriate value for $c$ can result in the selected subgoals being unable to effectively guide exploration, leading to the failure of the active exploration strategy. For all the experiments in Section 5.2, we set $c$ to 50 for Ant Fourrooms and Ant Maze(W-shape) tasks, and set $c$ to 20 for other tasks.

{\bf Subgoal selection radius $r_g$}: LESC sample a candidate subgoal set $\mathcal{C}$ within the radius ($r_g$) neighborhood of current state. An excessively large $r_g$ may result in the sampled subgoals being unreachable from current state, while too small $r_g$ could lead to the selected subgoal failing to guide exploration for faraway regions. We set $r_g$ to 20 for all the tasks. 
\section{Conclusion}
To tackle the challenging sparse-reward problems, we designed a measure of \textit{prospect} for subgoals and proposed an active exploration strategy LESC, which takes into account the \textit{prospect} and \textit{novelty} measures for subgoals. LESC images promising sparse pathes of landmarks by planning in the goal space. It then computes the prospect measure for subgoals based on the selected landmark. Unlike SAGA, which generates subgoals compatible with the low-level policy by adversarial learning, LESC samples subgoals in the vicinity of the current state to guild exploration. In addition, To balance the prospect measure and the novelty measure, we introduced a state-specific balance coefficient, that dynamically considers the importance of different measures by taking into account the reachability of the selected landmark. Experimental results demonstrated that LESC outperformes the state-of-the-art baselines. 
\newpage

\bibliographystyle{named}
\bibliography{ijcai24}

\end{document}